\documentclass{llncs}

\usepackage{amsmath}
\usepackage{amssymb}
\usepackage{amsfonts}
\usepackage{mathtools}
\usepackage{authblk}

\usepackage{graphicx}
\usepackage{booktabs}
\usepackage{hyperref}
\usepackage{cite}

\usepackage{multirow}
\usepackage{array}

\title{DobicVLM: Aligning Chest X-Ray Report Generation with Clinically-Grounded Programmatic Rewards via Group Relative Policy Optimization}

\author{Thanni Adewuyi\inst{1,2,3},
Angelica Obayi\inst{1,2},
Andem Aniekan\inst{1},
Samuel Okoko\inst{1,2},
Angel Ezendu\inst{1,2},
Ephraim Usani\inst{1},
Ademide Animasaun\inst{1,2},
Philip Chibundu\inst{1},
Christian Maurice\inst{1},
Mary Donald Essien\inst{1},
Oluwaseun Odunsi\inst{1,2},
Oluwasegun Oguntuase\inst{1,2},
Abiodun Adereni\inst{1,2}}

\authorrunning{T. Adewuyi et al.}
\institute{Dobic Health \and HelpMum Africa \and University of Ibadan \\
\email{medical@dobichealth.com}}

\begin{document}

\maketitle

\begin{abstract}
Medical imaging is a cornerstone of diagnostics, yet automated chest X-ray report generation struggles with structural adherence, anatomical completeness, and semantic faithfulness. We introduce DobicVLM, a vision-language model combining supervised fine-tuning on MedGemma-4B with Group Relative Policy Optimization (GRPO) and clinically-grounded programmatic rewards.

Our approach uses interpretable, rule-based reward components; structural verification, anatomical checklist, semantic similarity, and length constraints to enforce clinical standards without neural reward models. Trained on ~1,000 de-identified image-report pairs from a private clinical dataset (with ethics approval and compliance to local regulations), DobicVLM is evaluated via blinded expert review on 69 held-out cases.

DobicVLM outperforms Gemini 2.5 Flash across the majority of criteria, achieving the highest impression accuracy (27.2\%) and medical terminology (86.5\%) compared to both Gemini 2.5 Flash and MedGemma-4B baselines, with minor trade-offs in completeness and referrals. This demonstrates GRPO's value for transparent alignment in resource-limited settings.

\end{abstract}

\section{Introduction}

Medical imaging, particularly Chest X-Rays (CXRs), is central to global diagnostics but faces a critical disparity between imaging volume and radiologist availability, leading to burnout and reporting delays~\cite{moor2023foundation}. While Automated Radiology Report Generation (ARRG) aims to mitigate this, transitioning from experimental models to clinical utility remains challenging. General Vision-Language Models (VLMs) often fail to meet the high-stakes demands of radiology, such as rigid structural protocols (e.g., distinct ``Findings'' and ``Impression'' sections) and strict anatomical faithfulness~\cite{tu2024medgemini}. Standard Supervised Fine-Tuning (SFT) frequently struggles to enforce these constraints, often resulting in ``hallucinations'' or formatting failures despite learning statistical patterns~\cite{messmer2024review}.

To address these limitations, we introduce \textbf{DobicVLM}, a specialized framework that aligns chest X-ray reporting with clinical standards via a two-stage training process. First, we perform SFT on the MedGemma-4B architecture~\cite{sellergren2025medgemma} to establish baseline competency. Second, we refine the policy using \textbf{Group Relative Policy Optimization (GRPO)}~\cite{shao2024grpo}, a critic-free reinforcement learning algorithm. Unlike traditional Reinforcement Learning from Human Feedback (RLHF), which relies on opaque and often unstable neural reward models~\cite{ouyang2022training}, GRPO optimizes the policy by contrasting generated outputs against a \textbf{programmatic, interpretable reward function}.

Our reward design explicitly encodes verifiable clinical rules: (1) structural adherence via regex, (2) anatomical completeness via checklists, (3) semantic faithfulness, and (4) length constraints. This approach ensures transparency and auditability—critical requirements for clinical deployment—by prioritizing rule-based verification over learned approximations.

\subsection{Contributions}
\begin{itemize}
    \setlength\itemsep{0em} 
    \item \textbf{DobicVLM Framework}: A robust pipeline for fine-tuning MedGemma-4B on chest X-ray data using efficient low-rank adaptation (LoRA), establishing strong baseline performance on real-world clinical reports.
    \item \textbf{GRPO for Clinical Alignment}: One of the first applications of GRPO to radiology, achieving superior enforcement of structural and anatomical constraints compared to SFT alone without a separate critic model.
    \item \textbf{Verifiable Programmatic Rewards}: A novel, clinically-grounded reward function that prioritizes interpretable rules to penalize hallucinations and ensure EHR-compatible formatting.
\end{itemize}

We evaluate DobicVLM on real-world cases from a selected health facility, demonstrating improvements over Gemini 2.5 Flash and MedGemma-4B baselines in diagnostic impression accuracy and structural compliance.

\section{Related Work}

\textbf{Vision-Language Models in Medical Imaging.}
Recent progress in vision-language models (VLMs) has extended general-domain architectures to medical applications. Foundational work in open-source multimodal learning, such as OpenFlamingo~\cite{awadalla2023openflamingo} and Qwen-VL~\cite{bai2023qwen}, established the viability of interleaving image and text data for few-shot reasoning. In the medical domain, LLaVA-Med~\cite{li2023llava} proposed a curriculum-based alignment of biomedical images and text using GPT-4-generated instruction data. Subsequent specialized models, such as MedGemma~\cite{sellergren2025medgemma} and BiomedGPT~\cite{zhang2023biomedgpt}, build on robust language backbones with additional pretraining on medical image-text pairs. Other notable efforts include MediVLM~\cite{yan2025medivlm} and domain-specific adaptations of general VLMs like Med-Gemini~\cite{tu2024medgemini}. Despite these advances, generic or broadly tuned VLMs often fail to enforce radiology-specific requirements, such as structured section formatting and anatomical completeness.

\textbf{Automated Radiology Report Generation.}
Automated chest X-ray report generation has been extensively studied. Early works relied on hierarchical encoder-decoder models with reinforcement learning rewards based on n-gram metrics~\cite{liu2021r2gen, chen2021crossmodal}. More recent approaches incorporate retrieval-augmented generation~\cite{nooralahzadeh2023knowledge} or interactive iterative refinement~\cite{tanida2023interactive} to improve factual consistency. However, standard automated metrics (e.g., BLEU, CIDEr) often correlate poorly with clinical accuracy, necessitating clinically aware metrics like RadCliQ~\cite{yu2023evaluating}. While Vision-language foundation models show promise for end-to-end generation~\cite{moor2023foundation}, supervised fine-tuning alone frequently results in hallucinations or deviation from reporting protocols~\cite{messmer2024review}, underscoring the need for constraint-aware optimization.

\textbf{Reinforcement Learning in Language Models.}
Reinforcement learning from human feedback (RLHF) is standard for aligning models but suffers from training instability~\cite{ouyang2022training}. While Direct Preference Optimization (DPO)~\cite{rafailov2023direct} offers a stable, reward-model-free alternative, its application to medical multi-modal tasks is nascent~\cite{wang2024medicaldpo}. Critic-free alternatives like Group Relative Policy Optimization (GRPO)~\cite{shao2024grpo} have demonstrated strong results in mathematical reasoning by comparing group outputs. Our approach parallels concepts from Constitutional AI~\cite{bai2022constitutional}, which uses rule-based AI feedback to steer generation. In medical domains, RL has been applied to report generation~\cite{li2022reinforced, bassi2025mrg}, but typically relies on opaque neural reward models. We introduce programmatic, verifiable rewards with GRPO to ensure transparent clinical alignment.

\section{Methodology}

\subsection{Model Architecture and Training Infrastructure}

The model is built on MedGemma 4B, extending Gemma 3~\cite{gemma2025} for medical imaging. It includes a SigLIP 400M vision encoder~\cite{zhai2023siglip} that processes X-ray images $x \in \mathbb{R}^{H \times W \times C}$ into embeddings $\mathbf{v} = \mathcal{E}_v(x) \in \mathbb{R}^{N \times d}$. The language decoder supports image-text interleaving and 128K token context.

We use Low-Rank Adaptation (LoRA)~\cite{hu2021lora} for efficiency: for weight $W_0 \in \mathbb{R}^{d \times k}$, the forward pass is $h = W_0 x + \frac{\alpha}{r} BA x$, with $B \in \mathbb{R}^{d \times r}$, $A \in \mathbb{R}^{r \times k}$, and $r \ll \min(d,k)$. LoRA is applied to query and value projections in attention layers. Experiments run on Nvidia A100 GPUs with FP16 mixed-precision.

\subsection{Training Pipeline}

Training has two stages: Supervised Fine-Tuning (SFT) followed by Group Relative Policy Optimization (GRPO).

\subsubsection{Supervised Fine-Tuning (SFT)}

SFT adapts MedGemma on paired data $\mathcal{D}_{\text{SFT}} = \{(x_i, y_i)\}_{i=1}^N$, maximizing log-likelihood:
\begin{equation}
\mathcal{L}_{\text{SFT}} = -\mathbb{E}_{(x,y) \sim \mathcal{D}_{\text{SFT}}} \left[ \sum_{t=1}^T \log P_\theta(y^t \mid \mathbf{v}, p, y^{<t}) \right]
\end{equation}
Images are augmented with $\pm 5^\circ$ rotations, 0.3-probability flips, and [0.9, 1.1] contrast. Training: 5 epochs, lr $5 \times 10^{-5}$, AdamW with 500-step warmup. Checkpoint $\theta_{\text{SFT}}$ initializes GRPO.

\subsubsection{Group Relative Policy Optimization (GRPO)}

GRPO~\cite{shao2024grpo} refines the policy by generating $G$ candidates $\{y^{(1)}, \ldots, y^{(G)}\} \sim \pi_{\theta_{\text{old}}}(\cdot \mid x, p)$, assigning rewards $r_i = R(y^{(i)}, y^*, x)$, and computing advantages $A_i = \frac{r_i - \mu}{\sigma}$ (group mean/std).

Objective:
\begin{align}
\mathcal{L}_{\text{GRPO}}(\theta) &= \mathbb{E}_{x \sim \mathcal{D}} \Bigg[ \frac{1}{G} \sum_{i=1}^G \min \Bigg( \frac{\pi_\theta(y^{(i)} \mid x, p)}{\pi_{\theta_{\text{old}}}(y^{(i)} \mid x, p)} A_i, \notag \\
&\quad \text{clip}\left(\frac{\pi_\theta(y^{(i)} \mid x, p)}{\pi_{\theta_{\text{old}}}(y^{(i)} \mid x, p)}, 1-\varepsilon, 1+\varepsilon\right) A_i \Bigg) \Bigg] \notag \\
&\quad - \beta \mathbb{D}_{\text{KL}}(\pi_\theta \| \pi_{\text{ref}})
\end{align}
with $\mathbb{D}_{\mathrm{KL}}(\pi_\theta \,\|\, \pi_{\text{ref}}) = \mathbb{E} \left[ \log \frac{\pi_\theta}{\pi_{\text{ref}}} \right]$, $\pi_{\text{ref}} = \pi_{\theta_{\text{SFT}}}$.

Training: 1 epoch on $\mathcal{D}_{\text{SFT}}$ subset, lr $1 \times 10^{-5}$, $G=6$, $\varepsilon=0.2$, $\beta=0.01$.

\subsection{Clinically-Grounded Reward Function}

The reward $R(y, y^*, x)$ uses four programmatic components, prioritizing rule-based verification (ROUGE-L as auxiliary for similarity).

\textbf{Structural Adherence ($R_{\text{struct}}$):} Regex verification of headers (Indication, Comparison, Findings, Impression):
\begin{equation}
R_{\text{struct}}(y) = \begin{cases} 1.0 & \text{if correct} \\ 0.0 & \text{otherwise} \end{cases}
\end{equation}

\textbf{Anatomical Checklist ($R_{\text{check}}$):} Fraction of terms in the anatomical set $\mathcal{A}$ (containing terms such as cardiac silhouette, costophrenic angles, \dots, lung fields):
\begin{equation}
R_{\text{check}}(y) = \frac{|\mathcal{A} \cap \text{terms}(y)|}{|\mathcal{A}|}
\end{equation}

\textbf{Semantic Faithfulness ($R_{\text{overlap}}$):} ROUGE-L F1 on Impression sections:
\begin{equation}
R_{\text{overlap}}(y, y^*) = \text{ROUGE-L-F1}(\text{Impression}(y), \text{Impression}(y^*))
\end{equation}

\textbf{Length Constraint ($R_{\text{len}}$):} Soft penalty for $\ell(y)$ outside [80, 450]:
\begin{equation}
R_{\text{len}}(y) = \begin{cases} 1.0 & \ell_{\min} \leq \ell(y) \leq \ell_{\max} \\ \exp(-\gamma |\ell(y) - \text{clip}(\ell(y))|) & \text{otherwise} \end{cases}
\end{equation}
with $\gamma=0.01$.

Total: $R_{\text{total}} = \sum \lambda_i R_i$, $\lambda = [0.30, 0.30, 0.35, 0.05]$.

\section{Experiments}
\label{sec:experiments}

\subsection{Dataset and Experimental Setup}

\textbf{Training Data.} Supervised fine-tuning and GRPO refinement were performed on approximately 1,000 de-identified, high-quality chest X-ray report pairs sourced from a selected health facility. This dataset was curated to reflect real-world clinical reporting practices, with diverse pathologies and consistent adherence to institutional protocols. While the scale is modest reflecting practical constraints on accessing large volumes of annotated clinical imaging data due to privacy, cost, and availability, it enables effective domain-specific adaptation of the MedGemma-4B base model through efficient LoRA training.

\textbf{Evaluation Data Curation.} We evaluate on a separate curated set of 69 real-world chest X-ray cases from a selected health facility. The dataset represents diverse pathologies including normal findings, cardiomegaly, pneumonia, pleural effusions, and complex multi-finding cases. \textbf{While the sample size was constrained by the high resource intensity and financial cost of securing multiple board-certified radiologist reviews, the set was stratified to ensure coverage of both common and rare pathological presentations, an approach consistent with recent expert-curated radiology AI benchmarks that prioritize diagnostic challenge and clinical relevance over large-scale automated testing (e.g., the Radiology’s Last Exam (RadLE) benchmark, which uses only 50 spectrum-biased ``spot diagnosis'' cases to rigorously differentiate frontier models from human experts~\cite{datta2025radle}).} Each case was processed by all three models in a single forward pass (one-shot generation) without fine-tuning on test data.

\textbf{Models Evaluated.} We compare three state-of-the-art vision-language models:
\begin{itemize}
    \item \textbf{Gemini 2.5 Flash}: Google's latest multimodal model
    \item \textbf{MedGemma-4B}: Base medical VLM 
    \item \textbf{DobicVLM (Ours)}: MedGemma-4B + GRPO with verifiable rewards
\end{itemize}

\textbf{Blinded Expert Evaluation.} All generated reports were anonymized and randomized to eliminate bias. Four board-certified Nigerian physicians, including two radiologists and two medical doctors, independently evaluated each report against the corresponding standard radiologist reports across five clinically meaningful criteria using a standardized 5-point Likert scale:

\begin{enumerate}
    \item \textbf{Accuracy of Impression}: Does the clinical impression correctly identify primary findings?
    \item \textbf{Medical Terminology}: Appropriate use of clinical language and radiology conventions
    \item \textbf{Hallucination Rate}: Absence of fabricated findings not visible in the image
    \item \textbf{Report Completeness}: Adequate coverage of relevant anatomical structures
    \item \textbf{Appropriate Referral}: Correct recommendations for follow-up or additional imaging
\end{enumerate}

We computed mean Likert scale ratings for each criterion and model, aggregating across $n=276$ total evaluations (69 cases $\times$ 3 models $\times$ variable number of raters per case). 

\textbf{Rationale for Human Evaluation.} Unlike general image captioning, radiology report generation is a high-stakes clinical task where automated metrics like BLEU and ROUGE poorly correlate with diagnostic accuracy~\cite{miura2021improving}. A report can have high ROUGE-L overlap while containing critical diagnostic errors, or conversely, use different phrasing while being clinically equivalent. \textbf{Expert radiologist assessment remains the gold standard for validating clinical AI systems, as recently demonstrated in benchmarks such as Radiology’s Last Exam (RadLE)~\cite{datta2025radle}, which uses blinded expert scoring on a small set of challenging cases to reveal limitations in frontier multimodal models that automated metrics often miss.}

\subsection{Evaluation Metrics}

For each criterion $c$ and model $m$, we compute the mean Likert scale rating:
\begin{equation}
\text{Score}(m, c) = \frac{\sum_{i=1}^{N} \text{rating}_i \text{ for criterion } c}{N}
\end{equation}

\section{Results}
\label{sec:results}

\subsection{Overall Performance}

Table~\ref{tab:expert_eval} presents the expert evaluation results. DobicVLM establishes a distinct advantage over the generalist Gemini 2.5 Flash baseline across all metrics, validating the necessity of domain-specific fine-tuning.

Against the specialized MedGemma-4B baseline, our model achieves the highest performance on \textbf{Medical Terminology (86.5\%)} and the critical \textbf{Accuracy of Impression (27.2\%)} metric. While the improvement in impression accuracy over the MedGemma baseline is marginal (+0.9\%), it represents a qualitative refinement in identifying primary pathologies in complex cases.

Regarding safety, DobicVLM significantly reduces hallucinations compared to Gemini 2.5 Flash (65.1\% vs 41.0\% acceptance). However, we observe a slight regression compared to the Base MedGemma model (68.8\%). This suggests that while GRPO effectively steers the model toward clinically accurate impressions, the exploration inherent in reinforcement learning may occasionally introduce minor non-factual elaborations compared to the more conservative generation patterns of the base model.

\begin{table}[t]
\centering
\caption{Expert radiologist evaluation across 69 real-world chest X-ray cases (276 total evaluations). Percentages indicate acceptance rate. \textbf{Bold} indicates best performance per criterion.}
\label{tab:expert_eval}
\small
\begin{tabular}{lccc}
\toprule
\textbf{Criterion} & \textbf{Gemini 2.5} & \textbf{MedGemma} & \textbf{DobicVLM} \\
& \textbf{Flash} & \textbf{(Base)} & \textbf{(Ours)} \\
\midrule
Accuracy of Impression & 10.7\% & 26.3\% & \textbf{27.2\%} \\
Medical Terminology & 82.3\% & 83.2\% & \textbf{86.5\%} \\
Low Hallucination Rate & 41.0\% & \textbf{68.8\%} & 65.1\% \\
Report Completeness & \textbf{70.0\%} & 62.7\% & 60.2\% \\
Appropriate Referral & 36.1\% & \textbf{48.3\%} & 30.9\% \\
\bottomrule
\end{tabular}
\end{table}

\subsection{Clinical Significance}

DobicVLM's superior accuracy in clinical impressions--the most critical component of a radiology report--makes it more suitable for clinical deployment.

The trade-off in report completeness (60.2\% vs 70.0\% for Gemini) reflects our reward weighting: we intentionally prioritize \textit{accurate} reporting over \textit{exhaustive} reporting. This design choice reduces the model's tendency to generate verbose descriptions that mention all anatomy simply to appear complete---a behavior that contributed to Gemini's higher hallucination rate.
\subsection{Qualitative Analysis}

\begin{figure}[!ht]
    \centering
    \includegraphics[width=0.85\linewidth]{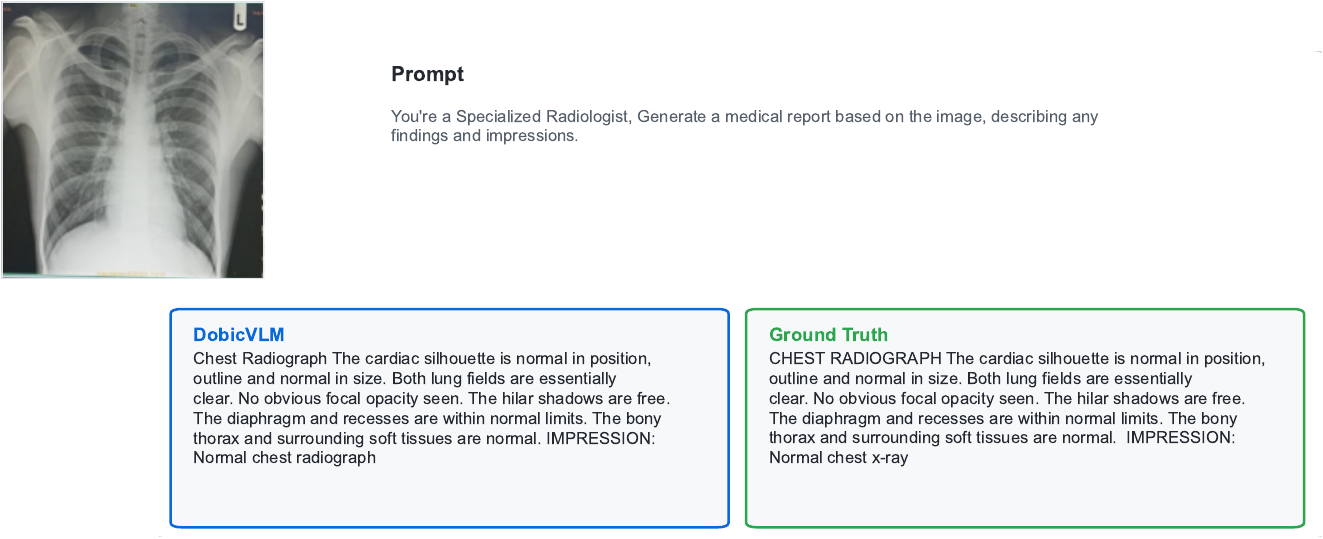} 
    \caption{Qualitative demonstration of DobicVLM. The model output demonstrates alignment with the ground truth radiologist impression, correctly identifying the normal cardiac silhouette and clear lung fields.}
    \label{fig:qualitative}
\end{figure}

Figure~\ref{fig:qualitative} presents a representative example where DobicVLM successfully identifies the clinical state of the patient. The generated report accurately reflects the visual evidence, maintaining high diagnostic accuracy and producing a structured impression that aligns with the radiologist's ground truth.

\section{Discussion}

\subsection{Clinical Implications and Strict Evaluation}

Our results show DobicVLM improves diagnostic accuracy while maintaining interpretability through programmatic rewards. The 2.5$\times$ gain in impression accuracy over Gemini 2.5 Flash (27.2\% vs 10.7\%) advances clinical AI usability.

Absolute rates require context: our strict evaluation rejected reports for minor omissions or inaccuracies in severity, unlike n-gram benchmarks. The 27.2\% acceptance highlights the gap to radiologist standards. DobicVLM suits draft generation with mandatory review or as a resident teaching tool.

\subsection{Mechanisms of Improvement and Trade-offs}

\textbf{Hallucination-Exploration Trade-off.} DobicVLM outperforms Gemini but shows slight hallucination regression vs. MedGemma (65.1\% vs 68.8\%). This stems from GRPO's exploration, favoring creative phrasings for semantic rewards over SFT's conservatism—potentially introducing unsupported details via "reward hacking."

\textbf{Reward Alignment and Referrals.} Referral scores dropped (30.9\% vs 48.3\%), likely because $R_{\text{overlap}}$ focuses on Impression, deprioritizing separate recommendation sections within length constraints.

\subsection{Limitations and Future Work}

\textbf{Metric Limitations.} ROUGE-L is an imperfect diagnostic proxy, favoring overlap over factuality. Future work: entity extraction (e.g., RadGraph~\cite{jain2021radgraph}) for pathology rewards.

\textbf{Dataset and Generalizability.} Limited to $\sim$1,000 training and 69 evaluation cases from one institution (Dobic Health Africa), constraining diversity. Inter-rater variability affects scores; data de-identified per regulations.

Automated generation needs human oversight. Next steps: (1) entity rewards for findings, (2) dedicated referral component, (3) multi-institution scaling and clinical trials.

\section{Conclusion}

We presented DobicVLM, combining SFT, GRPO, and programmatic rewards for CXR report alignment. Evaluation on 69 cases showed gains in impression accuracy (27.2\%) and terminology (86.5\%) over baselines, with reduced hallucinations vs. Gemini. This supports interpretable alignment in limited-resource settings. Limitations: modest scale; future: visual verification and validation.

\bibliographystyle{splncs04}
\bibliography{references}

\end{document}